\documentclass{article}
\usepackage{spconf}
\usepackage{amsmath,amssymb}
\usepackage{graphicx}
\usepackage{booktabs}
\usepackage{url}
\usepackage{cite}
\usepackage{microtype}

\title{FlexCast: Adaptive Weather Forecasting from Arbitrary Field Sets}

\name{%
\begin{tabular}{c}
Yuang Zhang$^{1}$\qquad
Chen Hui$^{1,\ast}$\qquad
Weisi Lin$^{2}$\quad
Haiqi Zhu$^{3}$\\ 
Xiulai Wang$^{1}$\qquad
Sun-Yuan Kung$^{4}$\qquad
Feng Jiang$^{1}$\qquad
\end{tabular}%
\thanks{$^{\ast}$Corresponding author:
\texttt{chui@nuist.edu.cn}.}%
}

\address{%
$^{1}$Nanjing University of Information Science and Technology \quad
$^{2}$Nanyang Technological University\\
$^{3}$Harbin Institute of Technology\quad
$^{4}$Princeton University
}
\usepackage{float}
\begin{document}
\ninept
\setlength{\abovedisplayskip}{1pt}
\setlength{\belowdisplayskip}{1pt}
\setlength{\abovedisplayshortskip}{0pt}
\setlength{\belowdisplayshortskip}{0.5pt}
\setlength{\jot}{0pt}
\makeatletter
\renewcommand{\section}{%
  \@startsection{section}{1}{\z@}%
    {-2.0ex plus -0.4ex minus -0.1ex}
    {0.5ex plus 0.05ex}
    {\normalfont\bfseries\centering}}
\renewcommand{\subsection}{%
  \@startsection{subsection}{2}{\z@}%
    {-1.2ex plus -0.3ex minus -0.1ex}
    {0.3ex plus 0.05ex}
    {\normalfont\bfseries}}
\makeatletter
\maketitle

\begin{abstract}
\vspace{-4pt}
Most deep learning weather models assign a fixed set of variables
and pressure levels to predefined channels, limiting transfer
across atmospheric field configurations. This dependence on a fixed field set limits the transferability of trained models across atmospheric field configurations. We propose FlexCast, a field-adaptive weather forecasting model that uses a single set of parameters to produce identity-aligned forecasts for variable-cardinality subsets drawn from a 69-field ERA5 registry. Specifically, a metadata-conditioned adapter the first encodes variable identity, pressure level, and field type and combines them with spatial features. Then, shared rank-16 projections are modulated by metadata-dependent gates to produce field-specific features, while masked set fusion aggregates the available fields into a fixed-width representation. Subsequently, a multiscale U-Transformer processes the fused atmospheric features, while an identity-aware query decoder produces forecasts for the requested fields. Finally, FlexCast learns a standardized six-hour increment and applies it recursively to generate forecasts at longer lead times. Experiments on the 2020 ERA5 test set demonstrate that FlexCast operates across varying field configurations. Compatible cross-field context is associated with lower forecast errors, whereas mismatched context increases them.
\end{abstract}

\vspace{-4pt}

\section{Introduction}
Deep learning-based weather forecasting has made substantial progress in medium-range prediction. Models based on Fourier operators, graph propagation, three-dimensional architectures, and cascaded forecasting have improved the representation of atmospheric dynamics\cite{fourcastnet,graphcast,pangu,fengwu,fuxi}. WeatherBench and WeatherBench 2 have also established common datasets and evaluation protocols\cite{weatherbench,weatherbench2}. However, most existing models are trained with a fixed collection of atmospheric variables and pressure levels, in which each field is assigned to a predefined channel. Such a formulation is effective when the field configuration remains unchanged, but limits the transferability of trained models when different combinations of atmospheric fields are available.

\begin{figure}[t]
  \centering
  \includegraphics[width=0.98\columnwidth]{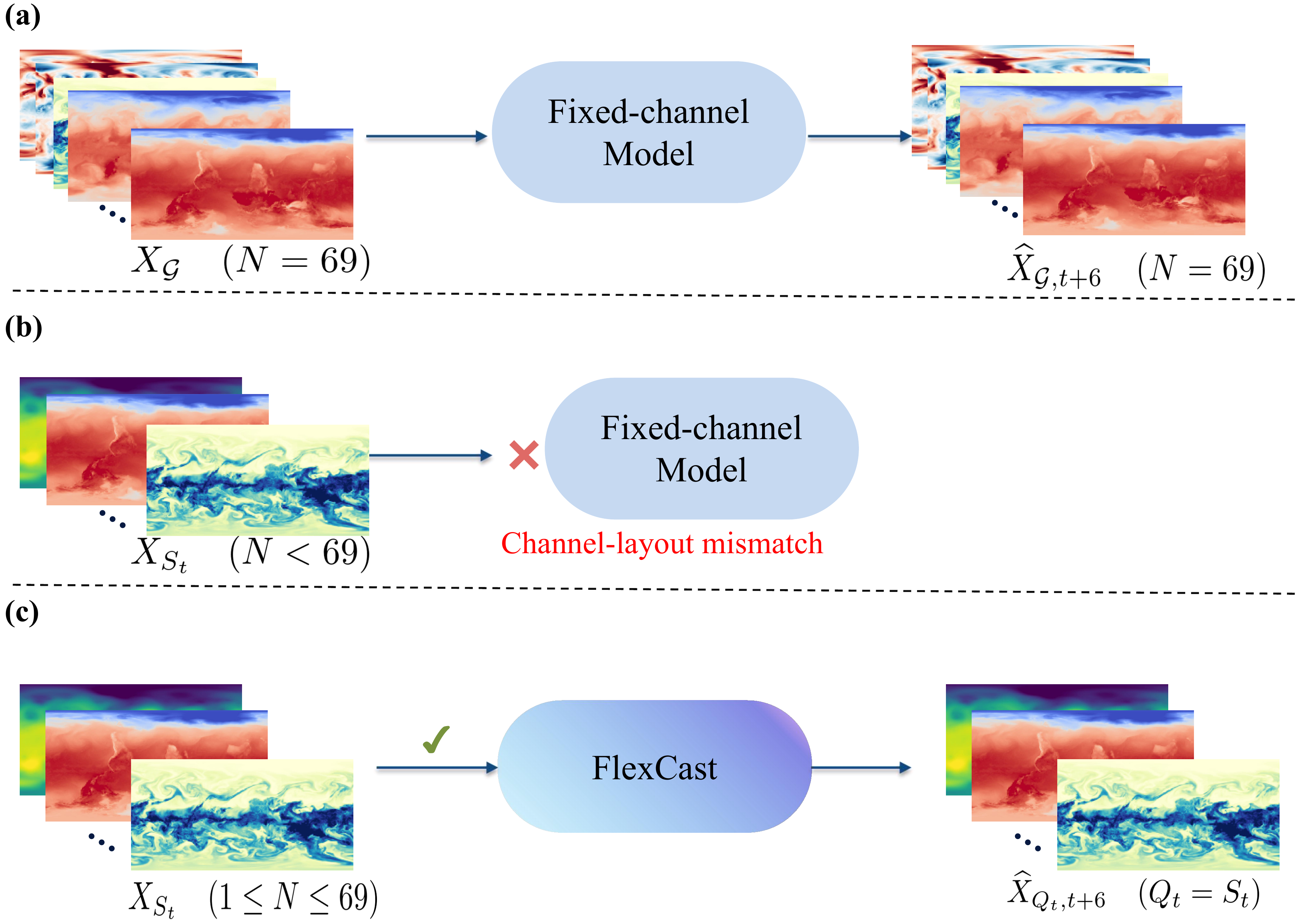}
  \caption{Comparison of fixed-channel forecasting and FlexCast for arbitrary atmospheric field sets. (a) A fixed-channel model forecasts the complete 69-field state. (b) A changed input subset is incompatible with the fixed-channel model’s predefined layout. (c) FlexCast forecasts the supplied fields from valid subsets with a single model. Here, $\mathcal{G}$ denotes the 69-field registry, $X_{\mathcal{G}}$ is the complete input, $S_t\subseteq\mathcal{G}$ is the available subset, $N=|S_t|$, and $Q_t=S_t$ are the requested forecast fields.}
  \label{fig:motivation}
\end{figure}

A change in the available field set affects more than the number of input channels. Temperature fields at 850 and 500 hPa describe the same meteorological variable at different atmospheric levels. Surface fields and pressure-level fields also have distinct physical identities. A model operating on arbitrary field sets must therefore represent the identity of each supplied field, combine information from a variable number of fields, and preserve the correspondence between its predictions and the requested outputs. Masking absent channels alone does not remove the dependence of conventional models on predefined channel positions. As illustrated in Fig. 1, the desired model accepts both the complete field collection and its valid subsets, while returning forecasts aligned with the identities of the supplied fields.

Recent studies have explored variable-adaptive expert allocation, heterogeneous atmospheric inputs, and shared representations for weather prediction \cite{vamoe,oneforecast,climax,aurora}. Convolutional, graph-based, spherical, and space-time formulations offer different ways to model atmospheric dynamics \cite{weyn2020,weyn2021,keisler,earthformer,sfno,neuralgcm,gencast}. Studies of scaling behavior and lightweight adaptation have further examined the roles of data coverage, model capacity, and parameter-efficient transfer \cite{scalinglawsweather,lehmann2026}. The setting considered here is more specific: the atmospheric field registry is fixed, but the subset supplied to the model can change between forecast cases, and that subset also defines the prediction targets. The objective is to learn a single forecasting operator for these arbitrary field configurations without training separate models or constructing field-specific forecasting branches.

To address this setting, FlexCast combines metadata-conditioned
rank-16 updates \cite{film,lora,hypernetworks} with masked Set
Fusion to obtain a fixed-width spatial representation. A shared
U-Transformer and identity-aware Query Decoder predict
standardized six-hour increments for the requested fields;
recursive application gives longer-lead forecasts.

Experiments on the 2020 ERA5 test set examine forecasting under different field cardinalities and cross-field context configurations. The results show that compatible context and broader atmospheric coverage generally improve forecast accuracy, whereas mismatched context increases prediction error. Overall, the main contributions of FlexCast are summarized as follows:
\begin{itemize}
  \item We propose a field-adaptive weather forecasting model, FlexCast, which relaxes the fixed-field assumption of conventional weather models and produces identity-aligned forecasts from variable-cardinality subsets of a 69-field ERA5 registry. The same parameter set is used across different field combinations, from sparse subsets to the full 69-field collection.
  \item In FlexCast, a metadata-conditioned adapter is designed to combine the spatial features of each field with its variable identity, pressure level, pressure value, and field type. Shared rank-16 projections and metadata-dependent gates then generate field-specific feature updates. Masked set fusion aggregates the available features into a fixed-width representation for the shared forecasting backbone.
  \item A multiscale U-Transformer is employed to model spatial dependencies in the fused atmospheric representation, while an identity-aware query decoder reconstructs forecasts for the requested fields. In addition, a standardized six-hour increment is learned and reused recursively, allowing the same model to produce forecasts at longer lead times.
\end{itemize}
\vspace{-6pt}
\begin{figure*}[t]
  \centering
  \includegraphics[width=\textwidth]{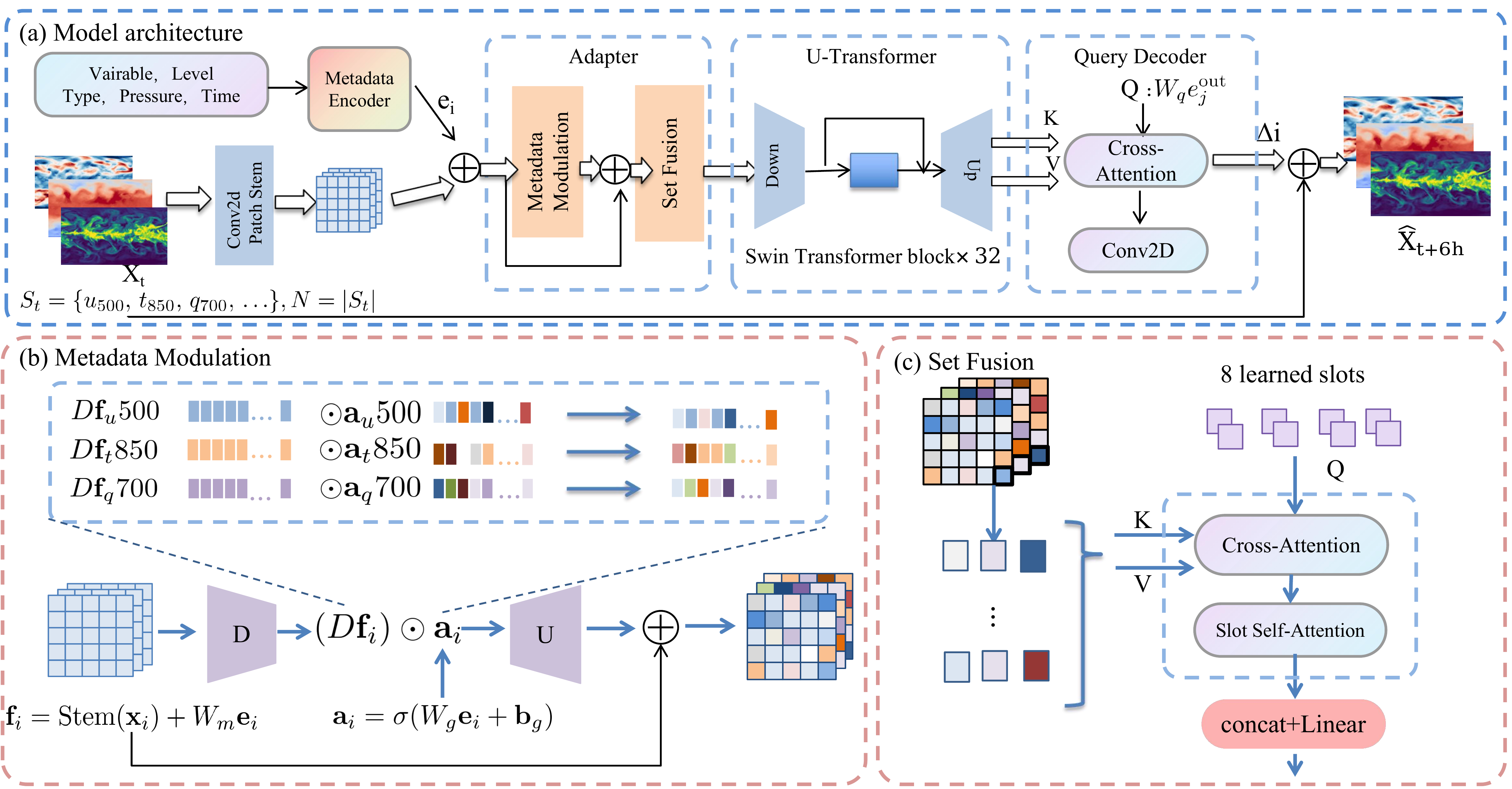}
\caption{Illustration of the proposed FlexCast framework.
(a) Overall forecasting pipeline. Each field in the available
set $S_t$ is encoded using its spatial map and metadata. The
adapted field features are fused and processed by a shared
U-Transformer, while identity-aware queries produce forecasts
for the requested set $Q_t=S_t$, with $N=|S_t|$. From left to right, the three addition nodes correspond to metadata conditioning, the gated low-rank feature update, and residual reconstruction using the de-standardized six-hour increment. (b) Metadata-conditioned
modulation, where the shared projections $D$ and $U$ reduce
and restore the feature width, and $\odot$ denotes element-wise
multiplication by the metadata gate. (c) Masked set fusion,
where learned slots attend to valid field tokens and are
refined by slot self-attention before projection to the spatial
representation.}
  \label{fig:architecture}
\end{figure*}

\section{Method}

\subsection{Overview of FlexCast}

Fig.~\ref{fig:architecture} illustrates the overall architecture of
FlexCast. The model first encodes each available atmospheric field
using its spatial map and physical metadata. The resulting
field-specific features are aggregated into a fixed-width spatial
representation, which is processed by a shared U-Transformer.
Identity-aware queries then decode forecasts for the requested fields. The same six-hour transition is recursively applied to obtain
longer forecast leads.

Let $\mathcal{G}=\{g_1,\ldots,g_{69}\}$ denote the predefined ERA5
field registry, containing five upper-air variables at 13 pressure
levels and four surface variables. At time $t$, the available subset
$S_t\subseteq\mathcal{G}$ contains $N=|S_t|$ fields, with
$1\leq N\leq69$. Each field $i\in S_t$ is represented by a normalized
spatial map $x_{i,t}\in\mathbb{R}^{H\times W}$ and metadata
$m_i=(v_i,\ell_i,p_i,c_i)$, describing its variable identity,
vertical-level identity, pressure in hPa, and field type. Given
$X_{S_t}=\{x_{i,t}\mid i\in S_t\}$ and field metadata $M_{S_t}$,
let $\mathcal{T}_t=(\tau_t,\tau_{t+6})$ collect the input- and
forecast-time features. FlexCast adopts same-set forecasting with
requested fields $Q_t=S_t$. The six-hour forecast is expressed as:
\begingroup
\setlength{\abovedisplayskip}{6pt}
\setlength{\abovedisplayshortskip}{6pt}
\setlength{\belowdisplayskip}{6pt}
\setlength{\belowdisplayshortskip}{6pt}
\begin{equation}
\widehat X_{Q_t,t+6}
=F_\theta(X_{S_t},M_{S_t},\mathcal{T}_t).
\label{eq:task}
\end{equation}
\endgroup
Each supplied field therefore defines a prediction target with the same identity.
The field cardinality, composition, and order may change between
samples, while the parameters $\theta$ remain shared. Field sets
within a batch are padded to a common size; padded entries are
excluded from attention and loss computation by a binary mask.

\subsection{Field-Adaptive Representation}

FlexCast first applies a metadata-conditioned adapter to the
available fields. Atmospheric fields may exhibit related spatial
patterns while representing different variables or vertical levels.
The adapter therefore combines shared spatial processing with
metadata-conditioned feature modulation. For field $i$, let
$E_i=E_v(v_i)+E_\ell(\ell_i)+E_c(c_i)$ denote the sum of learned
embeddings for its variable identity, vertical level, and field type.
For pressure-level fields, let $r_i=\log(p_i/1000)$ and
$P_i=g_p(\phi(r_i))$, where $\phi$ is a fixed Fourier encoding and
$g_p$ is a shared projection function. For surface fields, $P_i$
is set to zero. The time embedding is $T_\tau=g_t(\tau)$, where
$\tau$ contains the time features. The 64-dimensional metadata
embedding is defined as:
\begingroup
\setlength{\abovedisplayskip}{6pt}
\setlength{\abovedisplayshortskip}{6pt}
\setlength{\belowdisplayskip}{6pt}
\setlength{\belowdisplayshortskip}{6pt}
\begin{equation}
e_i=\operatorname{LN}(E_i+P_i+T_\tau).
\label{eq:metadata}
\end{equation}
\endgroup
Each field map is processed by the same $1\rightarrow96$ Conv2D
patch stem. With a $5\times4$ kernel, a $4\times4$ stride, and
latitude padding of 2, the stem maps a $121\times240$ input grid
to a $31\times60$ patch grid. The projected metadata is added
to every patch feature as
$f_{i,p}=f^{\mathrm{stem}}_{i,p}+W_m e_i$, where $p$ indexes the
spatial patches.

A 16-dimensional gate is generated from the metadata as
$a_i=\sigma(W_g e_i+b_g)$. Using shared projections
$D:\mathbb{R}^{96}\rightarrow\mathbb{R}^{16}$ and
$U:\mathbb{R}^{16}\rightarrow\mathbb{R}^{96}$, the adapted patch
feature is expressed as:
\begingroup
\setlength{\abovedisplayskip}{6pt}
\setlength{\abovedisplayshortskip}{6pt}
\setlength{\belowdisplayskip}{6pt}
\setlength{\belowdisplayshortskip}{6pt}
\begin{equation}
f'_{i,p}=f_{i,p}+U\!\left((Df_{i,p})\odot a_i\right).
\label{eq:modulation}
\end{equation}
\endgroup
The projections $D$ and $U$ are shared by all registry fields,
whereas $a_i$ depends on field metadata. Thus, the effective
field-dependent update $U\operatorname{diag}(a_i)D$ has rank at
most 16 and does not require a separate adapter for each field
\cite{hypernetworks,film,lora}.

The adapted field features are then combined by masked set fusion.
At each spatial patch, these features form a set of $N$ tokens.
FlexCast uses eight learned inducing slots to aggregate this
variable-cardinality set into a fixed-width spatial representation
\cite{settransformer,perceiverio}. At patch $p$, let $F_p$ denote
the normalized field-token matrix, $R_p^{(k)}$ the slot matrix in
fusion block $k$, and $Q_p^{(k)}=\operatorname{LN}(R_p^{(k)})$.
In each of the two fusion blocks, the masked multi-head
cross-attention update \cite{attention} is written as:
\begingroup
\setlength{\abovedisplayskip}{6pt}
\setlength{\abovedisplayshortskip}{6pt}
\setlength{\belowdisplayskip}{6pt}
\setlength{\belowdisplayshortskip}{6pt}
\begin{equation}
\bar R_p^{(k)}
=R_p^{(k)}+\operatorname{MHA}(Q_p^{(k)},F_p,F_p;B_M).
\label{eq:set_attention}
\end{equation}
\endgroup
Here, $B_M$ masks padded field positions in the attention logits.
Slot self-attention and an MLP refine the updated slots after each
cross-attention operation. The final eight slots are flattened and
projected to a 192-channel spatial token, while the slot features are
retained for output decoding. Set-based processing of valid field
tokens makes the fused representation invariant to their input order.

\subsection{Multiscale Forecasting and Identity-Aware Decoding}

The fused tokens are then arranged into a
$31\times60\times192$ feature map, which is processed by the
multiscale U-Transformer shown in Fig.~\ref{fig:architecture}.
The encoder changes the feature size from
$31\times60\times192$ to $16\times30\times384$ and then to
$8\times15\times768$. A symmetric decoder restores the spatial
resolution using skip connections. Patch Merge and Patch
Expansion change the resolution and channel width, while
shifted-window attention models spatial interactions at
different scales. Cyclic longitude padding maintains continuity
across the longitudinal boundary \cite{fuxi,unet,swin,swinv2}.
Since the backbone receives a fixed-width representation, its
parameters do not depend on the number or order of available
fields.

The shared forecast representation is decoded for the requested
fields. A fixed output-channel layout restricts the model to a
predefined field configuration. FlexCast instead generates a query
for each requested field $j\in Q_t$. Its field identity and
forecast-time features $\tau_{t+6}$ are encoded in
$e_j^{\mathrm{out}}$, yielding $q_j=W_qe_j^{\mathrm{out}}$.
At patch $p$, the context $C_p$ contains the projected backbone
token and retained relation slots. Let
$A_{j,p}=\operatorname{MHA}(q_j,C_p,C_p)$ denote the query
attention response. The query feature combines with the input
feature of the same field as:
\begingroup
\setlength{\abovedisplayskip}{3pt}
\setlength{\abovedisplayshortskip}{3pt}
\setlength{\belowdisplayskip}{3pt}
\setlength{\belowdisplayshortskip}{3pt}
\begin{equation}
h_{j,p}=q_j+A_{j,p}+W_{\mathrm{match}}f'_{j,p}.
\label{eq:query}
\end{equation}
\endgroup
The query features are arranged on the patch grid and mapped
to the original $121\times240$ resolution by a shared
transposed-convolution head. This head produces the standardized increment
map $\widehat d_j$ for each requested field. For a batch of
size $B$, the output has shape $[B,N,121,240]$ and follows the
requested field order. Together with order-invariant set fusion,
the query-based decoder ensures that reordering the supplied
fields only reorders their corresponding predictions.
\subsection{Loss Function and Training Strategy}

FlexCast predicts a standardized six-hour increment rather
than the next atmospheric state directly. Let $\mu_j^\Delta$
and $\sigma_j^\Delta$ denote the mean and standard deviation
of the six-hour increment for field $j$, estimated from the
training period. For the normalized field maps defined above,
the standardized target increment and reconstructed forecast
are expressed as:
\begin{equation}
\begin{aligned}
d_j
&=\frac{x_{j,t+6}-x_{j,t}-\mu_j^\Delta}
{\sigma_j^\Delta},\\
\widehat{x}_{j,t+6}
&=x_{j,t}+\mu_j^\Delta+\sigma_j^\Delta\widehat d_j .
\end{aligned}
\label{eq:residual}
\end{equation}

To account for the area represented by each grid cell and the
relative importance of different fields, the model is trained
with a masked, latitude-weighted, field-weighted mean-squared
error. For sample $b$ and field $j$, the per-field error and
overall objective are formulated as:
\begin{equation}
\begin{aligned}
\ell_{bj}
&=\frac{1}{HW}\sum_{h=1}^{H}\sum_{w=1}^{W}
a_h(\widehat d_{bjhw}-d_{bjhw})^2,\\
\mathcal{L}
&=\frac{\sum_{b,j}m_{bj}\omega_j\ell_{bj}}
{\sum_{b,j}m_{bj}\omega_j},
\end{aligned}
\label{eq:loss}
\end{equation}
where $a_h$ is the normalized cosine-latitude weight,
$\omega_j$ is the weight assigned to field $j$, and $m_{bj}$
indicates whether the field is valid in sample $b$. Padded
fields therefore do not contribute to the loss.

For a forecast horizon of $k$ six-hour steps, FlexCast recursively
applies the same forecasting operator at each step. Each predicted
state is used as the input to the next step, while the subset of
field identities remains fixed. During training, forecast horizons
and complete, dense-subset, and random-subset field configurations
are sampled. The loss in Eq.~\ref{eq:loss} is evaluated on the
valid fields at each supervised forecast target.
\vspace{-5pt}
\section{Experiments}
\subsection{Experimental Setup}

We use six-hourly ERA5 data at $1.5^\circ$ resolution on a
$121\times240$ grid \cite{era5}. The model is trained on data
from 1979--2018, while the 2019 data are used for checkpoint
selection. The 2020 test set provides 1460 valid forecast
starts for the short-lead experiments. FlexCast is trained
on eight NVIDIA A100 GPUs, and the same preprocessing and
optimization settings are used for its different field
configurations.

Forecast accuracy is measured by latitude-weighted RMSE
after field-wise normalization, macro-averaged over the
scored fields (nRMSE). All FlexCast results use the best checkpoint. The supplied fields also define the
requested outputs, following the same-set rule $Q_t=S_t$.
For the long-range Full-69 evaluation, the 4-, 7-, and
10-day results are computed from 1424 starts with complete
forecast windows. For comparison, FengWu, Pangu-Weather,
and FuXi are re-evaluated using the same ERA5 preprocessing
and nRMSE evaluation protocol as FlexCast.

\subsection{Overall Forecasting Performance}

Table~\ref{tab:rq1} compares FlexCast with three global
forecasting models at five lead times. The FlexCast forecasts
are generated by recursively applying the same six-hour
model.
\vspace{-8pt}
\begin{table}[H]
\centering
\caption{Multi-lead forecasting results on the 2020 ERA5
test set. Values are nRMSE (lower is better) under
the same evaluation protocol. The best result in each
column is shown in bold, and the second best is underlined.}
\vspace{4pt}
\label{tab:rq1}
\small
\setlength{\tabcolsep}{2pt}
\begin{tabular}{@{}lccccc@{}}
\toprule
Model & 6 h & 1 day & 4 days & 7 days & 10 days \\
\midrule
FengWu
& \underline{0.1252} & \underline{0.2028}
& 0.4860 & 0.6887 & 0.8143 \\
Pangu-Weather
& \textbf{0.1182} & \textbf{0.1859}
& \underline{0.4509} & 0.6960
& \underline{0.7582} \\
FuXi
& 0.1258 & 0.2050 & 0.4742
& \underline{0.6484} & 0.7831 \\
FlexCast(Full-69)
& 0.1421 & 0.2269 & \textbf{0.4246}
& \textbf{0.5626} & \textbf{0.6341} \\
\bottomrule
\end{tabular}
\end{table}
\vspace{-6pt}
Table~\ref{tab:rq1} shows that Pangu-Weather achieves the lowest nRMSE at 6 h and 1 day. At longer
lead times, FlexCast gives the lowest errors among the
compared models, reaching 0.4246, 0.5626, and 0.6341
at 4, 7, and 10 days, respectively. Thus, although
FlexCast is less accurate at short leads, it obtains
lower nRMSE at longer lead times under the matched
evaluation protocol.

\subsection{Effects of Cross-Field Context}

To examine the use of cross-field information, we evaluate
the forecast of an anchor field while changing the auxiliary
fields supplied with it.The 13 anchor fields are $t_{850}$, $t_{500}$, $u_{850}$,
$u_{500}$, $v_{850}$, $q_{850}$, $q_{700}$, $z_{500}$,
$z_{850}$, $u_{10}$, $v_{10}$, $t_{2m}$, and MSL.
Here, $u_{10}$ and $v_{10}$ denote the 10-m wind components,
$t_{2m}$ denotes 2-m temperature, and MSL denotes mean
sea-level pressure. Same-variable context supplies other pressure levels of the anchor variable, whereas same-level context supplies other variables at the anchor’s pressure level.
These two conditions apply to the nine pressure-level
anchors. Anchor-only and broad-context conditions cover
all 13 anchors.

To further examine whether the auxiliary fields provide
compatible information, we replace their maps with maps from a sample displaced by 30 days while keeping the anchor and requested
target unchanged. All conditions use the same 2020 test
starts. Relative changes are computed for each anchor
before averaging.
\vspace{-10pt}
\begin{table}[H]
\centering
\caption{Cross-field context comparison at 6 h on the
2020 ERA5 test set. Values are mean anchor-field nRMSE
(lower is better). Relative changes are computed for each anchor and then averaged: the same-variable, same-level, and broad-matched conditions are compared with the corresponding anchor-only condition, whereas the broad-mismatched condition is compared with the broad-matched condition.}
\vspace{4pt}
\label{tab:context}
\small
\setlength{\tabcolsep}{3pt}
\begin{tabular}{@{}lccc@{}}
\toprule
Condition & Anchors & nRMSE & Change \\
\midrule
Anchor only & 13 & 0.1943 & -- \\
+Same-variable context & 9 & 0.1695 & $-10.2\%$ \\
+Same-level context & 9 & 0.1408 & $-21.0\%$ \\
+Broad matched context & 13 & 0.1269 & $-34.3\%$ \\
+Broad mismatched context & 13 & 0.2818 & $+124.5\%$ \\
\bottomrule
\end{tabular}
\end{table}
\vspace{-3pt}

Compared with the corresponding anchor-only forecasts,
same-variable and same-level context reduce the mean
relative error by 10.2\% and 21.0\%, respectively.
Broad matched context produces a 34.3\% reduction across
all 13 anchors. In contrast, mismatched auxiliary maps
increase the mean relative error by 124.5\% compared
with broad matched context. It should be noted that the
nine-anchor and 13-anchor rows contain different targets;
their aggregate nRMSE values are therefore not directly
comparable. The results show that the value of additional
fields depends on the compatibility of the atmospheric
information they provide.

\subsection{Effects of Field-Set Cardinality}

We evaluate the same FlexCast model with different
numbers of supplied fields, without configuration-specific
fine-tuning. Table~\ref{tab:cardinality} presents the results
for five field-set cardinalities at 6, 12, 18,and 24h.
\vspace{-16pt}
\begin{table}[H]
\centering
\caption{Forecasting results for different field-set
cardinalities on the 2020 ERA5 test set. $N$ denotes the
number of available fields. Values are nRMSE under
same-set forecasting (lower is better). The best result is shown in bold, and the second best is
underlined.}
\vspace{3pt}
\label{tab:cardinality}
\small
\setlength{\tabcolsep}{4pt}
\begin{tabular}{@{}lrrrr@{}}
\toprule
$N$ & 6 h & 12 h & 18 h & 24 h \\
\midrule
8  & 0.1627 & 0.2232 & 0.2659 & 0.3007 \\
16 & 0.1662 & 0.2186 & 0.2539 & 0.2822 \\
32 & 0.1570 & 0.2032 & 0.2331 & 0.2563 \\
48 & \underline{0.1566} & \underline{0.2003}
   & \underline{0.2278} & \underline{0.2489} \\
69 & \textbf{0.1421} & \textbf{0.1997}
   & \textbf{0.2163} & \textbf{0.2269} \\
\bottomrule
\end{tabular}
\end{table}
\vspace{-3pt}

As shown in Table~\ref{tab:cardinality}, the complete
69-field configuration gives the lowest nRMSE at every
evaluated lead time. Compared with the eight-field subset,
its error decreases from 0.1627 to 0.1421 at 6 h and
from 0.3007 to 0.2269 at 24 h. However, the improvement
is not strictly monotonic: the eight-field subset slightly
outperforms the 16-field subset at 6 h. This indicates
that forecast accuracy depends not only on the number of
available fields but also on their identities and physical
information.

\subsection{Ablation Study}

To evaluate the contributions of the field-adaptive
components, we train two ablated variants of FlexCast.
One removes the metadata addition to the patch features,
and the other removes the modulation in
Eq.~\ref{eq:modulation}. Table~\ref{tab:ablation}
compares the retrained variants with the full model
at five forecast lead times.

\vspace{-9pt}
\begin{table}[H]
\centering
\caption{Ablation study of the metadata addition and modulation in FlexCast. Mean nRMSE (lower is better) is reported
across different lead times. The best result in each column
is shown in bold, and the second best is underlined.
``w/o'' denotes ``without''.}
\vspace{4pt}
\label{tab:ablation}
\small
\setlength{\tabcolsep}{1pt}
\begin{tabular}{@{}lccccc@{}}
\toprule
Method & 6 h & 1 day & 4 days & 7 days & 10 days \\
\midrule
FlexCast (Full-69)
& \textbf{0.1421} & \textbf{0.2269}
& \textbf{0.4246} & \textbf{0.5626}
& \textbf{0.6341} \\
w/o Metadata Addition
& 0.2420 & 0.4906 & 0.9178
& 1.1734 & 1.3965 \\
w/o Modulation
& \underline{0.1972} & \underline{0.3844}
& \underline{0.6798} & \underline{0.7641}
& \underline{0.8072} \\
\bottomrule
\end{tabular}
\end{table}
\vspace{-4pt}

Table~\ref{tab:ablation} shows that both ablated variants have higher nRMSE than the full model at every lead time. Without metadata addition, the error increases from
0.1421 to 0.2420 at 6 h and from 0.6341 to 1.3965 at
10 days. Removing the modulation also increases the
error, reaching 0.1972 at 6 h and 0.8072 at 10 days.
These results demonstrate the importance of both components, with metadata addition having the larger effect on forecast accuracy.

\section{Conclusion}
We propose FlexCast, a field-adaptive weather forecasting model that uses a single parameter set to generate identity-aligned forecasts for variable-cardinality subsets of a fixed 69-field ERA5 registry.
FlexCast uses a metadata-conditioned Adapter with shared rank-16 modulation and masked Set Fusion to represent and aggregate the available fields. A multiscale U-Transformer and an identity-aware Query Decoder then generate forecasts for the supplied fields. On the 2020 test set, FlexCast
achieves the lowest nRMSE among the compared models at 4, 7,
and 10 days. Compatible context improves accuracy, whereas
mismatched context degrades it. The full field set yields the
lowest error among tested configurations; ablations support
metadata addition and the modulation.
 {\fontsize{9.5}{9}\selectfont
\bibliographystyle{IEEEbib}
\bibliography{refs}
}

\end{document}